# GAN based ball screw drive picture database enlargement for failure classification

Tobias Schlagenhauf, Chenwei Sun, Jürgen Fleischer, Karlsruhe Institute of Technology

## ABSTRACT

The lack of reliable large datasets is one of the biggest difficulties of using modern machine learning methods in the field of failure detection in the manufacturing industry. In order to develop the function of failure classification for ball screw surface, sufficient image data of surface failures is necessary. When training a neural network model based on a small dataset, the trained model may lack the generalization ability and may perform poorly in practice. The main goal of this paper is to generate synthetic images based on the generative adversarial network (GAN) to enlarge the image dataset of ball screw surface failures. Pitting failure and rust failure are two possible failure types on ball screw surface chosen in this paper to represent the surface failure classes. The quality and diversity of generated images are evaluated afterwards using qualitative methods including expert observation, t-SNE visualization and the quantitative method of FID score. To verify whether the GAN based generated images can increase failure classification performance, the real image dataset was augmented and replaced by GAN based generated images to do the classification task. The authors successfully created GAN based images of ball screw surface failures which showed positive effect on classification test performance.

## 1. INTRODUCTION

In the manufacturing industry, monitoring the condition of the machine and its components is one of the core tasks to protect the machine, improve the quality of the production and optimize the operating cost [1]. For machines driven by the ball screw drive unit, the condition of the ball screw should be monitored so as to keep the machine running normally, repair or change the ball screw parts timely and extend the life of the machine. Thus, an automatic inspection function which can classify different kinds of surface failures of the ball screw and can be integrated in existing condition monitoring applications is desired. To build and train a classification model, sufficient, balanced and large variations of data are very important to guarantee its generalization performance and reduce its overfitting tendency during model training [5]. However, image data of ball screw surface failures is often limited, costly and time-consuming to generate. To address this issue, GAN as a method of generating image data based on a small number of preexisting images can be investigated.

Previously, modifying the images with affine transformations like rotation, cropping, flipping or with other effects including blur, sharpen, white balance tuning etc. has become the most-used methods of data augmentation [17]. In 2014, Generative Adversarial Network (GAN) [7] has been proposed and has been evidenced that it can be used in the field of image generation to generate synthetic images from random noise [7]. It can either be an alternative method to enrich the image dataset or it can be used together with classical data augmentation methods as a hybrid data augmentation method [6]. In this paper, GAN is implemented on images of ball screw surface failures and its effect on the classification performance is evaluated. The rest of the paper is organized as follows: Section 2 introduces state-of-the-art related work of GAN in data augmentation. Section 3 introduces the processes of GAN image generation, image evaluation and image classification with convolutional neural networks (CNN). In section 4, the results with respect to the quality of generated images and the classification results are presented. The conclusion is presented in section 5.

## 2. RELATED WORKS

According to [4], a standard GAN consists of two parts, the generator and the discriminator. The generator generates images from random noise (randomly sampled latent variables) while the discriminator differentiates whether its input images are real or generated. The goal of the discriminator is to minimize the error of its classification result while the generator tends to maximize this error, leading the discriminator to recognize the generated input images as real images. The left image of Figure 1 shows the concept of deep convolutional GAN (DCGAN). According to [7], the objective function for DCGAN is:

$$\min_G \max_D V(D,G) = E_{x \sim p_{data}(x)}[\log D(x)] + E_{z \sim p_z(z)}\left[\log\left(1 - D(G(z))\right)\right] \quad (1)$$

Here $p_z(z)$ is the distribution of a latent variable as generator input, in which values are randomly selected. The distribution of $p_z(z)$ is usually Gaussian. $p_{data}(x)$ is the distribution of the real data and $D(x)$ indicates the probability that image $x$ is classified as a real image.

The process of training a DCGAN is separated in two steps and is illustrated on the right side of Figure 1. First, the generator is frozen, and the discriminator is trained and is updated. Second, the discriminator is fixed, and the generator is trained with the knowledge of the updated discriminator parameters. Details are described as below:

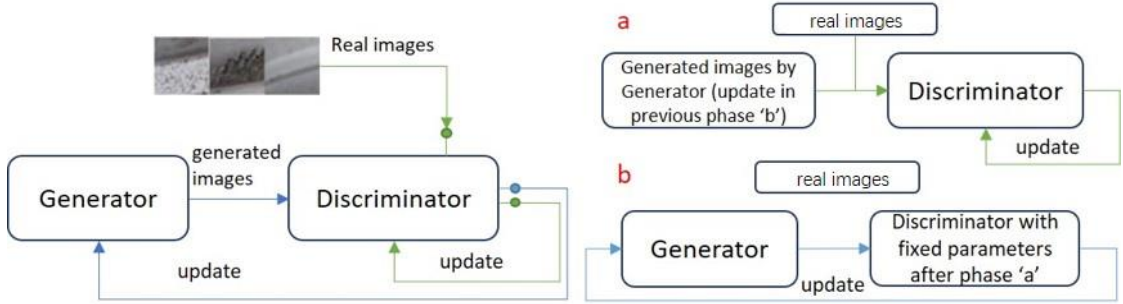

Figure 1 Left: Basic concept of DCGAN. Right: Illustration of weights update. a: train discriminator, back propagation starts from the output of discriminator to the input of discriminator. b: train generator, back propagation starts from the last layer of generator based on the output of fixed discriminator

1. Generator is fixed, discriminator is concerned at first. The goal is to maximize $V(D)$:

$$\max_D V(D) = E_{x \sim p_{data}(x)}[\log D(x)] + E_{z \sim p_z(z)}\left[\log\left(1 - D(G(z))\right)\right] \quad (2)$$

If real images are correctly classified as real images, the first part of the function becomes large, if generated images are correctly classified as well, $D(G(z))$ should be very small, which makes the second part of the function larger.

2. Discriminator is fixed, generator is concerned. The goal is to minimize $V(G)$:

$$\min_G V(G) = E_{z \sim p_z(z)}\left[\log\left(1 - D(G(z))\right)\right] \approx -E_{z \sim p_z(z)}\left[\log\left(D(G(z))\right)\right] \quad (3)$$

Training the generator will go through the entire GAN and all weights and biases are updated except the part that belongs to the discriminator. If generated images are incorrectly classified as real images, $D(G(z))$ should be large, which makes $V(G)$ small.

The minimization of $V(G)$ with the optimum $D_G^*(x)$ in the phase of training the discriminator is described as:

$$\min_G V(G,D) = C(G) = -2\log 2 + 2JSD\left(p_{data}(x) || p_g(x)\right) \quad (4)$$

Apparently, the index called Jenson Shannon Divergence $JSD\left(p_{data}(x) || p_g(x)\right)$ [7] should be minimized to achieve the goals of minimizing $C(G)$. In other words, $JSD$ can be used to describe the similarity of two distributions, the smaller the JSD, the more similar are the real data to the generated data. The training process goes on until the generated distribution from generator is close to the distribution of original dataset (example see in Figure 2). In this paper, both the generator and discriminator use convolutional neural network, where some architecture design experiences were derived from [16].

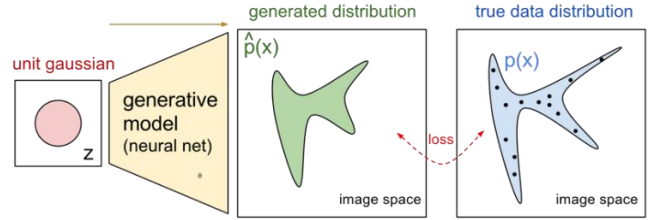

Figure 2 Generator takes unit gaussian random noise points as input and output a distribution in image space which imitates the true data distribution [14]

Recently research showed different approaches of using GANs to generate images as the source of data augmentation. [1] tested a data augmentation GAN which uses a U-net and Resnet combined generator structure and a WGAN discriminator to enhance the vanilla classifier and one-shot learning. [3] used GAN to generate synthetic images for medical image segmentation and did tests to compare the data augmentation effects of only using GAN, using GAN and rotation augmentation combined and using different ratios of generated images. The result showed that GAN can provide a great improvement for image segmentation performance. [13] used a so-called BAGAN and an autoencoder-based initialization to augment minority class images for an imbalanced dataset to get better image quality and achieve a good classification result in comparison with other GANs for data augmentation. In the field of surface failure classification, [19] did metallic surface failure detection and enlarge the failure image dataset using classical data augmentation. [2] used CycleGan to augment both failure and non-failure data and get better result in comparison with the accuracy without any data augmentation. [11] did sharpness data augmentation of steel surface defect images and use YOLO to do real time defect classification. To conclude, GAN based data augmentation showed great potential to enlarge small datasets and showed positive effects on improving the classification performance.

## 3. Method

To generate synthetic images with GAN and apply them as the source data for ball screw surface failure classification, we created a five-step methods: firstly, the original image dataset is collected and each class in the whole dataset is divided into a train and a test subset. The train set is used for GAN image generation and for training the classification model. The test set is used for evaluating the classification performance. Secondly, classical data augmentation methods are used to enlarge the failure image train sets. Thirdly, DCGAN is implemented on surface failure images. After that, the quality of the DCGAN generated images is evaluated. Lastly, several classification comparative tests are done to evaluate the influence of GAN generated images on the classification test performance. Details are described in subsections below.

3.1 Data source

The original ball screw dataset consisted of 40 images taken from a worn Ball Screw Drive. To collect features of ball screw surface failures, two possible failures on the ball screw: pitting and rust were selected and cropped from the complete worn ball screw drive.

The area of surface failure was far less than that of intact ball screw surface, which means that the number of cropped images of pitting and rust was also far less than that of intact surface. In order to get these data, two different cropping methods were implemented: manual cropping for the surface failure images and programmed automatic cropping for intact surfaces. Figure 3 shows how the pitting images were cropped. Because the pitting failure area was not regulated, some parts might be very small, while some were very wide and large. In consideration that the input images of the GAN and of the CNN classification model should be resized to the same size, uniform width and height of the cropped failure images was aspired to avoid a large deformation during rescaling. Especially for long pitting failure, it was better to crop only a part of it. In this paper, all cropped images were resized to 96 x 96 pixels.

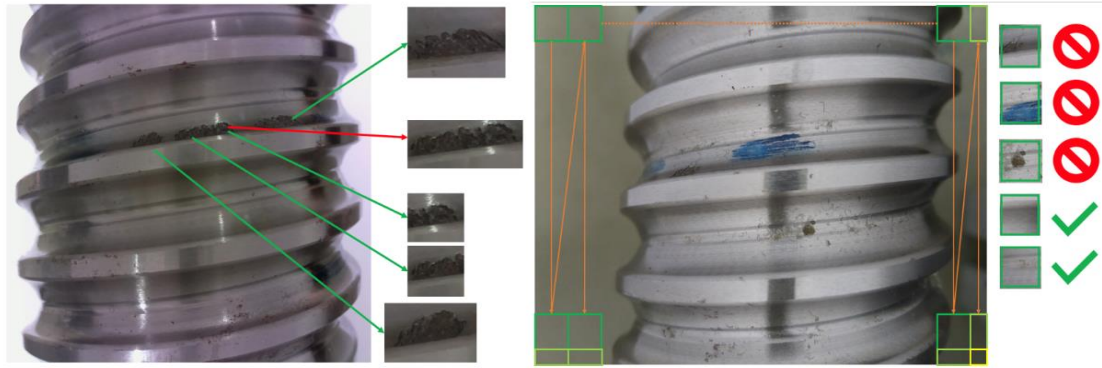

Figure 3 cropping methods of surface failure and intact surface parts. Left: cropping of the pitting images. Red arrow points out the unsuitable cropping dimension. Right: filtering the automatic cropped intact surface images. The cropping examples of the first three on the right side which do not belong to intact surface should not be included.

The cropped images were then separated in train and test set, and the number of them are listed in Table 1:

Table 1 Number of original cropped images for each class

| Classes | Total Number | Train Number | Test Number |
| --- | --- | --- | --- |
| Pitting | 132 | 100 | 32 |
| Rust | 169 | 128 | 41 |
| Intact | 922 | 700 | 222 |

3.2 Classical data augmentation

Classical data augmentation has been proved to be an effective method of enlarging the dataset and reducing the possibility of overfitting [15]. Some typical data augmentation approaches include affine transformation, noise addition, brightness modifying, detail enhancing, nonlinear deformation etc. (Figure 4). In this paper, flipping, rotation, translation and additional gaussian noise were chosen as classical data augmentation methods.

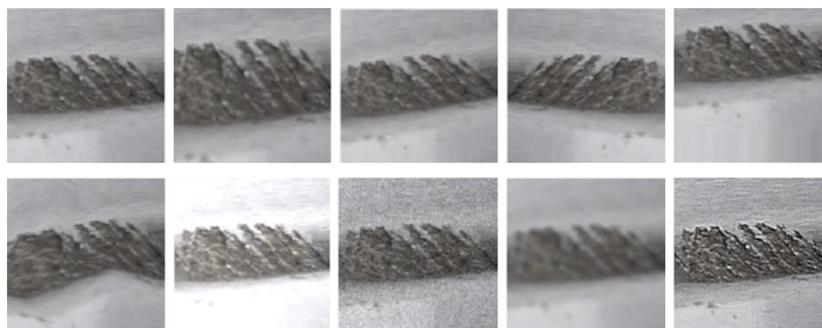

Figure 4 Classical data augmentation examples: the image on the top left is the original one, the images to the right from top to bottom are images after amplifying, rotation, flipping, translation, elastic deformation, brightness tuning, adding gaussian noise, use gaussian filter and edge enhancement

3.3 GAN implementation and corresponding evaluation

In this paper, all codes were programmed with Python 3.6. All models were built with Keras backend on TensorFlow and were run on a local machine with GPU RTX2070 Max-Q. GAN was implemented on the train set of original pitting images, train set of original rust images, classical augmented train set of pitting images and classical augmented train set of rust images respectively. The structures of the generator and the discriminator are listed in Table 2:

Table 2 Architectures of generator and discriminator

|    | **Generator** | **Discriminator** |
|----|---------------|-------------------|
| 1  | Latent vector (100-2000) | Gaussian Noise Layer (optional) |
| 2  | Dense + Reshape: 6x6 @1024 | Conv Layer: 96x96 @32 with 5x5 filter |
| 3  | Transposed Conv Layer: 12x12 @512 with with 4x4 filter | LeakyReLU Activation + Dropout 25% |
| 4  | BN Layer + ReLU Activation | Conv Layer: 48x48 @64 with 5x5 filter |
| 5  | Transposed Conv Layer: 24x24 @256 with 4x4 filter | BN + LeakyReLU + Dropout 25% |
| 6  | BN Layer + ReLU Activation | Conv Layer: 24x24 @128 with 3x3 filter |
| 7  | Transposed Conv Layer: 48x48 @128 with 4x4 filter | BN + LeakyReLU + Dropout 25% |
| 8  | BN Layer + ReLU Activation | Conv Layer: 12x12 @256 with 3x3 filter |
| 9  | Transposed Conv Layer: 96x96x3 (1 if gray) with 4x4 filter | BN + LeakyReLU + Dropout 25% |
| 10 | | Flatten Layer |
| 11 | | Dense Layer: 3 (1 if gray) |

In this paper, the quality of GAN generated surface failure images was evaluated qualitatively and quantitatively. Two methods were used as qualitative evaluation: 1. Experts were asked to identify real images in a mixture of generated and real images. 2. t-SNE dimension reduction method [12] were used to estimate whether the distribution of generated image dataset coincided with real image dataset. To quantitatively evaluate the quality, Fréchet Inception Distance (FID) [10] was used. The lower the FID value, the more similar the distribution of generated image dataset to the original image dataset is. The authors also used latent space interpolations to verify whether the GAN learned the generalized features of the training image dataset or it just remembered the features and tended to overfitting. If the generated images based on interpolated latent variables showed strong transition change effects, the generalized features were learned.

Several tests with different hyperparameters were conducted to choose the best series to generate images with good generated image quality based mainly on visualized image quality and FID score. In each test, GAN was trained for 10000 loops where in each loop one batch of 50 real images were randomly selected and used in the discriminator. According to test results in Figure 5, 50 images as batch size, latent vector size of 100, 4 times discriminator training per loop (k_d), 3 times generator training per loop (k_g), Adam [9] optimizer with a learning rate 0.0002 and $\beta_1 = 0.5$ were chosen. Additionally, adding a Gaussian noise layer was helpful to prevent non-intersection between the low dimensional input latent space and the high dimensional original images [16].The one-sided label smoothing method [18] proved to have a good effect on avoiding the gradient vanishing [16] especially when training with rust images, although the quality of the generates images evaluated by the FID score showed a slight reduction. Additional tests using the number of training loops as variable showed that FID scores decreased until 25000 loops, the loss curve of the generator and the discriminator stays almost unchanged and the change of visualized images quality and FID scores was negligible. In Table 3, the selected hyperparameters are listed.

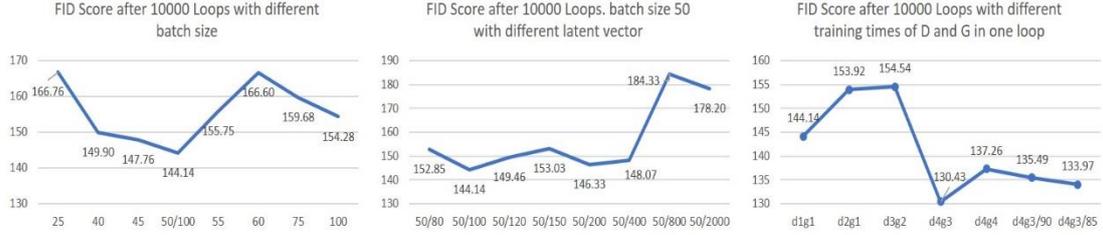

Figure 5 Hyperparameter selection. 50/100 in the left and middle graphs means batch size is 50, latent vector size is 100. In the right graph, all tests are based on using batch size 50 and latent vector size 100. d4g3 means discriminator trains 4 times, generator trains 3 times in one training loop. "/85" means the extra one-sided label smoothing method (when training discriminator, the real image label is set as 0.85, while the fake image label is set as 1-0.85=0.15).

Table 3 Training parameters of DCGAN

| Classes | Batch Size | Training Loops | k_d | k_g | latent vector size | Optimizer Discriminator | Optimizer Generator | Discriminator with Gaussian Noise Layer | Label for real in Discriminator | Label for fake in Discriminator |
|---|---|---|---|---|---|---|---|---|---|---|
| Pitting | 50 | 30000 | 4 | 3 | 100 | Adam (0.0002,0.5) | Adam (0.0002,0.5) | yes | 1 | 0 |
| Rust | 50 | 30000 | 4 | 3 | 100 | Adam (0.0002,0.5) | Adam (0.0002,0.5) | yes | 0.85 | 0.15 |

3.4 Classification and evaluation

The goal of the classification in this paper is to verify the effects of GAN based generated images on improving the classification test performance. Six classification tasks were designed:

1. Classification with original small and imbalanced image dataset.

2. Classification with data augmentation to balance the datasets of failure classes using GAN generated images.

3. Classification with GAN generated images to replace the entire images in failure train classes.

4. Classification with classical data augmentation to balance the datasets of failure classes with the methods mentioned in 3.2.

5. Classification with data augmentation using GAN generated images. GAN was trained with original images and classical augmented images. Each failure class had been augmented to 700 images.

6. Classification with classical data augmentation which based on the original images and 100 GAN generated pitting images per failure class.

The number of train and test sets of the original dataset (from Table 1) in classification 1 and the number of each class in classification 2 to 6 are listed in table 4:

Table 4 Number of each class in classification tasks

| Classes | original | | after data augmentation/ replacement of train set | |
|---|---|---|---|---|
| | train | test | train | test |
| Pitting | 100 | 32 | 700 | 32 |
| Rust | 128 | 41 | 700 | 41 |
| Intact | 700 | 222 | 700 | 222 |

The test sets of the failure images were always independent from any classical data augmentation and GAN training and were only used for classification testing. For all classification tasks, a simple 3-layer convolutional neural network was built as below:

Table 5 Classification network architecture

|   | **Classification Network** |
|---|---|
| 1 | Conv Layer: 96x96 @16 with 5x5 filter |
| 2 | BN + ReLU + MaxPooling |
| 3 | Conv Layer: 48x48 @32 with 5x5 filter |
| 4 | BN + ReLU + MaxPooling |
| 5 | Conv Layer: 24x24 @64 with 3x3 filter |
| 6 | BN + ReLU + MaxPooling |
| 7 | Flatten Layer |
| 8 | Dense Layer: 64 with ReLU |
| 9 | Dense Layer: 3 with Softmax activation function (label is one-hot encoded) |

The evaluations of classification results were done on the test set. Because the number of images in each class was not equal, using simple average accuracy was not suitable to evaluate the classification performance. Hence, other metrics were considered: recall, precision, F1 score and macro average, which computes the metrics above independently for each class and then outputs the average [4, 8].

To simplify all processes above, a GUI for GAN training and image classification (Figure 6) was designed with the following purposes:

1. Visualization of the results much better;

2. Simplification of the program execution with only blanks filling and buttons clicking;

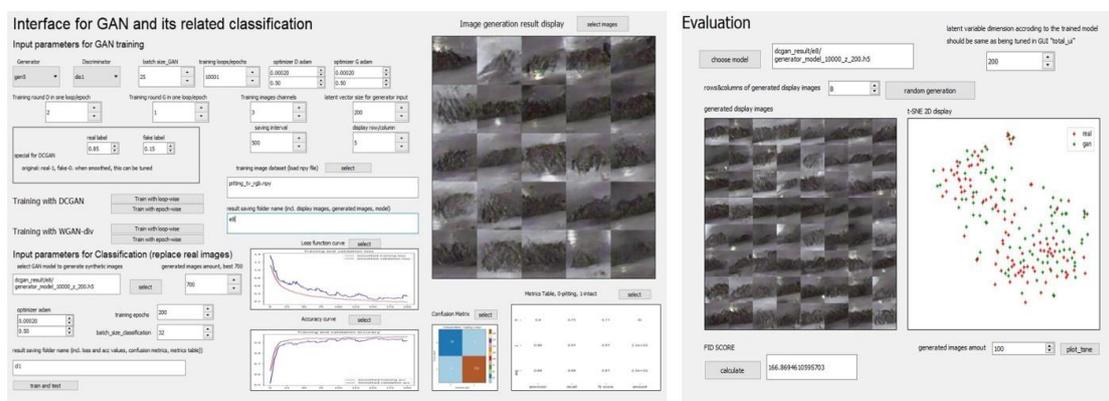

Figure 6 GUI design for GAN and following classification tasks

## 4. RESULTS & DISCUSSION

4.1 Generated Image Samples

The generated pitting and rust images are shown in Figure 7 and in Figure 8 respectively. The generator models were used in following evaluation and classification tasks.

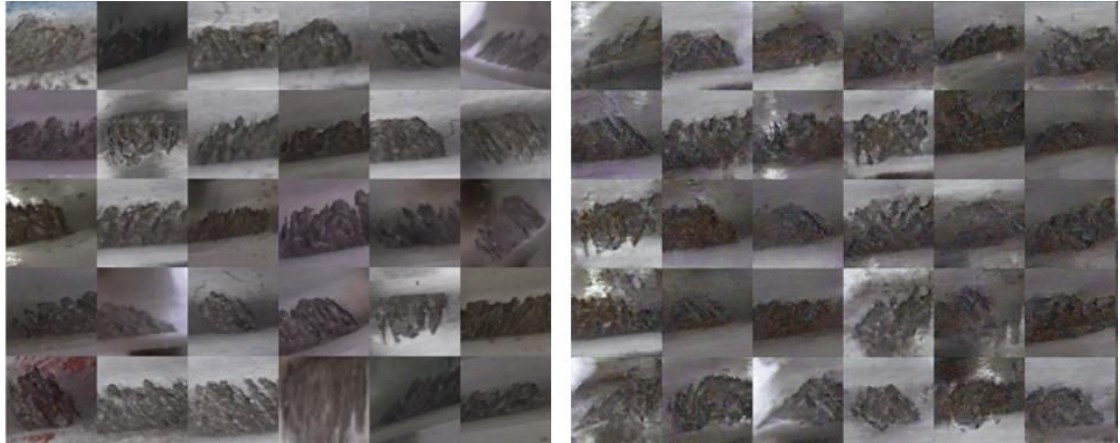
Figure 7 Left: real pitting images. Right: GAN generated pitting images

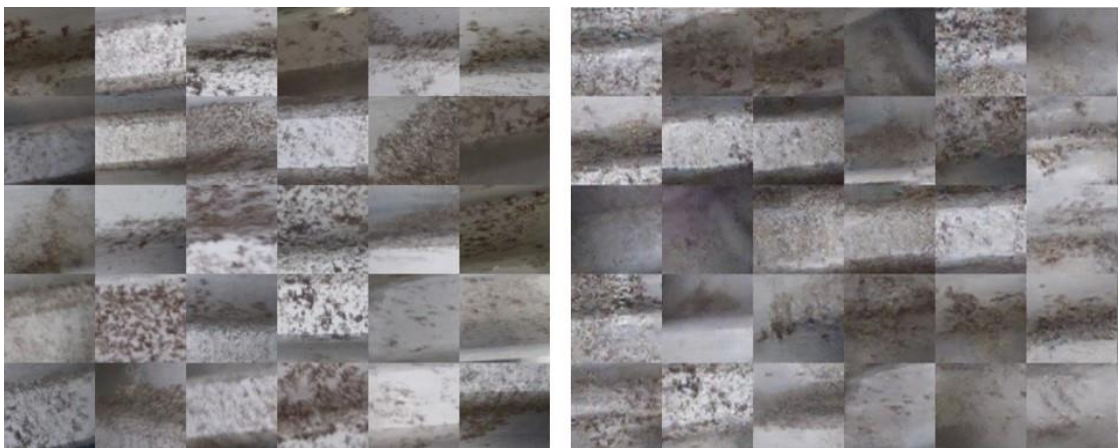
Figure 8 Left: real rust images. Right: GAN generated rust images

4.2 Evaluation Result

To calculate FID score, the trained generator model was used to generate an image dataset where the number of images in the dataset is the same as in the original image dataset. Because the FID scores of pitting and rust images were different when using different generated images, they were calculated 10 times to get the average value respectively. The FID scores are listed in Table 6.

Table 6 FID Score

|  | Compare with original data | Compare with classical augmented data |
|---|---|---|
| Pitting | 110.2 | 108.1 |
| Rust | 98.5 | 98.6 |

To verify whether the trained GAN learned the features of the real images, the interpolation of the latent vector was done to generate the transition images between 2 selected real images. Figure 9 shows the transition effects. Theoretically, the number of transition images can be infinity. The transition images provide additional information to the original dataset and can be used to enlarge the dataset.

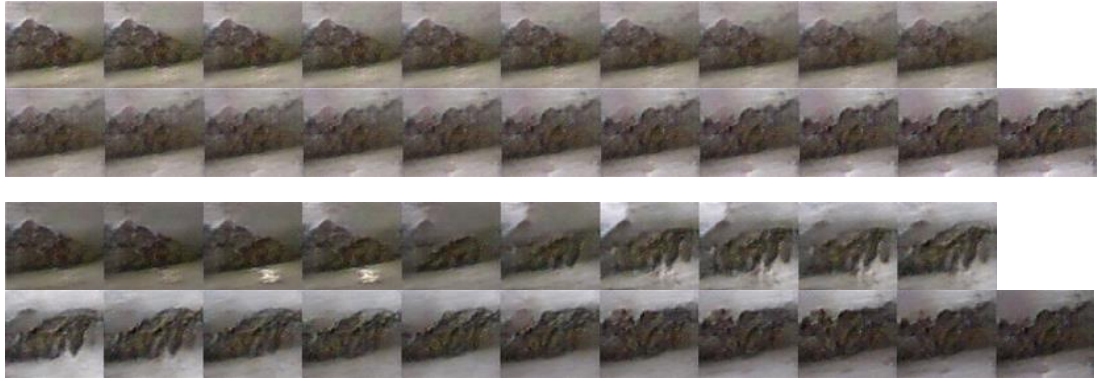

Figure 9 Top: Interpolation of two images at the two sides of the image row, the transparency ratio of two images are changed. Bottom: Interpolation of latent vector of two images at each side of the image row. The transition images between them are generated which are based on the interpolated latent vector and can be viewed clearly.

Additionally, t-SNE was used to visualize and compare the similarity of the distributions of images by projecting them in 2-dimensional space. Figure 10 shows the similarity of the GAN generated image data distribution to the distribution of the original image data. Additionally, it indicated that a higher number of generated images led to a more continuous distribution of the discrete original images.

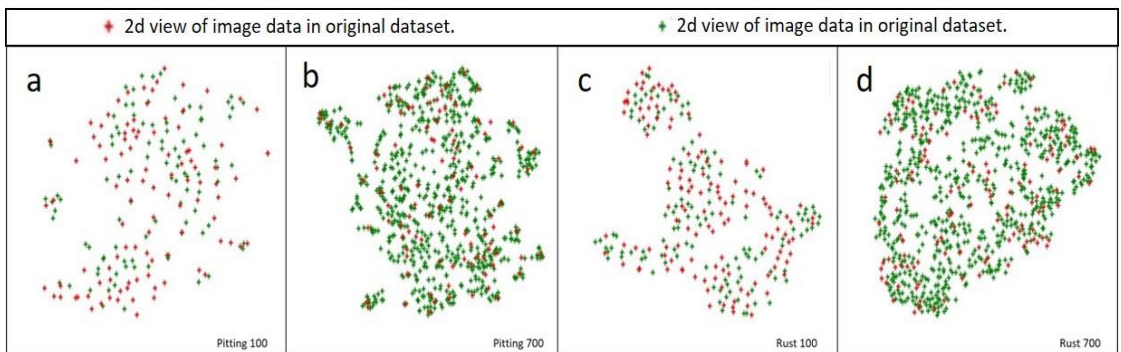

Figure 10 t-SNE visualization of the distribution comparison in 2d space a: comparison of image dataset of real pitting images and generated pitting images dataset with the same number of images in each class. b: comparison of image dataset of real pitting images and 700 generated pitting images. c: comparison of image dataset of real rust images and generated rust images with the same number of images in each class. d: comparison of image dataset of real rust images and 700 generated rust images

4.3 Classification Result

Table 7 lists the hyperparameters for training the classification network:

Table 7 Hyperparameters of training the classification model

| Classes | Batch Size | Training epochs | Optimizer | Label of pitting | Label of intact | Label of rust |
|---|---|---|---|---|---|---|
| Pitting | 32 | 100 | Adam (0.0002,0.5) | 0 | 1 | 2 |

An example to show the result of the 3. Classification task is illustrated in Figure 11.

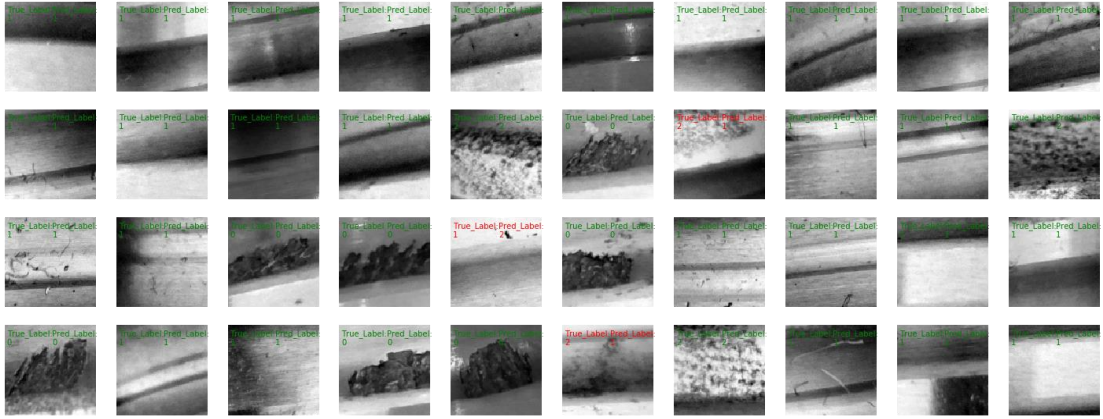

Figure 11 Correct and wrong classification samples from the test image set

The result of all classification tasks described in 3.5 are listed in Table 8. Each setup was classified 10 times to compute averaged metrics.

Table 8 Classification results

|  | 1. Classification (original) | | | | 2. Classification (original+GAN) | | | | 3. Classification (GAN) | | | |
| --- | --- | --- | --- | --- | --- | --- | --- | --- | --- | --- | --- | --- |
|  | precision | recall | F1-score | Overall Accuracy | precision | recall | F1-score | Overall Accuracy | precision | recall | F1-score | Overall Accuracy |
| 0-pitting | 0.923 | 0.816 | 0.867 | 0.909 | 0.921 | 0.836 | 0.88 | 0.9371 | 0.957 | 0.814 | 0.887 | 0.9343 |
| 1-intact | 0.957 | 0.966 | 0.96 | | 0.976 | 0.967 | 0.971 | | 0.966 | 0.973 | 0.971 | |
| 2-rust | 0.746 | 0.718 | 0.707 | | 0.747 | 0.81 | 0.787 | | 0.741 | 0.798 | 0.767 | |
| Macro average | 0.879 | 0.844 | 0.849 | | 0.881 | 0.88 | 0.88 | | 0.889 | 0.863 | 0.874 | |
|  | 4. Classification (original+classical) | | | | 5. Classification (original -classical-GAN) | | | | 6. Classification (original-GAN-classical) | | | |
|  | precision | recall | F1-score | Overall Accuracy | precision | recall | F1-score | Overall Accuracy | precision | recall | F1-score | Overall Accuracy |
| 0-pitting | 0.864 | 0.896 | 0.877 | 0.9328 | 0.856 | 0.949 | 0.899 | 0.9428 | 0.897 | 0.911 | 0.909 | 0.9457 |
| 1-intact | 0.986 | 0.96 | 0.969 | | 0.98 | 0.977 | 0.976 | | 0.999 | 0.954 | 0.972 | |
| 2-rust | 0.797 | 0.819 | 0.773 | | 0.819 | 0.767 | 0.787 | | 0.756 | 0.904 | 0.823 | |
| Macro average | 0.869 | 0.89 | 0.873 | | 0.886 | 0.896 | 0.89 | | 0.881 | 0.929 | 0.904 | |

Table 9 shows the changes of metrics based on classification using only original images (1. Classification). Two of all metrics were more significant: F1 score was the most generalized index to evaluate the classification result on the test set. All augmentation and replacement tests showed increasing trends, especially on F1 score of rust test images. In 4. Classification, F1 scores of all classes were higher than that of 2. Classification, which indicates that data augmentation with GAN generated images had better effect than that with classical augmented images. Both tests with combined methods (5. Classification and 6. Classification) had even higher test performance than that with only one implemented augmentation method. F1 scores of the 3. Classification showed great potential of training the classification model with fully GAN generated failure images.

Besides, the misclassification of failure images as intact surface should be avoided. The precision value of intact surface class is an important index which reflects this phenomenon. All tests of 2 to 6 showed increasing trends with the ratio of 1.99%, 0.94%, 3.03%, 2.40% and 4.39% respectively. It showed that during data augmentation or data replacement, the misclassifications of surface failure images as intact surfaces were suppressed, although classical data augmentation methods were simpler and more efficient.

Table 9 Classification results comparison based on 1. Classification

|  | 1. Classification (original) | | | | 2. Classification (original+GAN) | | | | 3. Classification (GAN) | | | |
| --- | --- | --- | --- | --- | --- | --- | --- | --- | --- | --- | --- | --- |
|  | precision | recall | F1-score | Overall Accuracy | precision | recall | F1-score | Overall Accuracy | precision | recall | F1-score | Overall Accuracy |

| | precision | recall | F1-score | Overall Accuracy | precision | recall | F1-score | Overall Accuracy | precision | recall | F1-score | Overall Accuracy |
|---|---|---|---|---|---|---|---|---|---|---|---|---|
| 0-pitting | 0.923 | 0.816 | 0.867 | | -0.22% | 2.45% | 1.50% | | 3.68% | -0.25% | 2.31% | |
| 1-intact | 0.957 | 0.966 | 0.96 | 0.909 | 1.99% | 0.10% | 1.15% | 3.09% | 0.94% | 0.72% | 1.15% | 2.78% |
| 2-rust | 0.746 | 0.718 | 0.707 | | 0.13% | 12.81% | 11.32% | | -0.67% | 11.14% | 8.49% | |
| Macro average | 0.879 | 0.844 | 0.849 | | 0.23% | 4.27% | 3.65% | | 1.14% | 2.25% | 2.94% | |
| | 4. Classification (original+classical) | | | | 5. Classification (original -classical-GAN) | | | | 6. Classification (original-GAN-classical) | | | |
| | precision | recall | F1-score | Overall Accuracy | precision | recall | F1-score | Overall Accuracy | precision | recall | F1-score | Overall Accuracy |
| 0-pitting | -6.39% | 9.80% | 1.15% | | -7.26% | 16.30% | 3.69% | | -2.82% | 11.64% | 4.84% | |
| 1-intact | 3.03% | -0.62% | 0.94% | 2.62% | 2.40% | 1.14% | 1.67% | 3.72% | 4.39% | -1.24% | 1.25% | 4.04% |
| 2-rust | 6.84% | 14.07% | 9.34% | | 9.79% | 6.82% | 11.32% | | 1.34% | 25.91% | 16.41% | |
| Macro average | -1.14% | 5.45% | 2.83% | | 0.80% | 6.16% | 4.83% | | 0.23% | 10.07% | 6.48% | |

## 5. CONCLUSION

In this paper, the authors used DCGAN to generate synthetic pitting and rust images to enlarge the original dataset of possible ball screw drive surface defects and evaluated the generated images using qualitative and quantitative methods. Subsequently, the authors used the images to do classification tasks in two ways: data augmentation and data replacement. In comparison with the classification result using the original image data, both GAN based data augmentation and classical data augmentation showed improvements with increases in the evaluation metrics on the test set. The GAN based data augmentation showed better classification results than the classic data augmentation approach when F1 scores and overall accuracy were used as metrics. The best results were achieved by combining the two data augmentation methods where classical data augmentation was applied on the original and GAN generated images (6. Classification task). Besides, it was evidenced that replacing all original real images in the train set with GAN generated images could train a model with good test performance (3. Classification task) as well.

GAN can be viewed as a more powerful data augmentation way to yield more generalized models. In this paper, GAN data augmentation has been found to be especially powerful when it was combined with classical data augmentation techniques. However, the improvement effects were not astonishing large for this setup. One possible reason for this might be the relatively small diversity in the training and test data which should be approached in further works together with testing the GAN approach against other classical data augmentation techniques.

## 6. ABBREVIATIONS

GAN – Generative Adversarial Network

CNN – Convolutional Neural Network

FID – Fréchet Inception Distance

GUI – Graphical User Interface

## 7. ACKNOWLEDGEMENTS